\documentclass[3p,twocolumn, round, sort & compress]{elsarticle}

\usepackage{hyperref}
\usepackage{graphicx}
\usepackage{times}
\usepackage{epsfig}
\usepackage{graphicx}
\usepackage{amsmath,cases,amssymb}
\usepackage{mathrsfs}
\usepackage{algorithm}
\usepackage{caption}
\usepackage{algorithmic}
\usepackage{bm}
\usepackage{multirow}
\usepackage{mathrsfs}
\usepackage{wrapfig}
\usepackage{booktabs}
\usepackage{subcaption}
\usepackage[switch]{lineno}
\usepackage{setspace}
\usepackage{color}
\usepackage{adjustbox}
\usepackage{mathtools}
\usepackage{dblfnote}
\usepackage{comment}
\usepackage{float}
\usepackage{lipsum}

\usepackage{url}
\modulolinenumbers[1]

\newif\ifshowfig\showfigtrue

\definecolor{violet}{rgb}{0.56, 0.0, 1.0}
\definecolor{cerulean}{rgb}{0.0, 0.48, 0.65}

\biboptions{authoryear}

\begin{document}

\begin{frontmatter}

\title{Conditional Local Feature Encoding for Graph Neural Networks}
\cortext[mycorrespondingauthor]{Corresponding author}
\author{Yongze Wang}
\ead{Yongze.Wang@student.uts.edu.au}

\author{Haimin Zhang}
\ead{Haimin.Zhang@uts.edu.au}

\author{Qiang Wu}
\ead{Qiang.Wu@uts.edu.au}

\author{Min Xu\corref{mycorrespondingauthor} }
\ead{Min.Xu@uts.edu.au}

\address{School of Electrical and Data Engineering, Faculty of Engineering and IT, University of Technology Sydney }
\address{15 Broadway, Ultimo, NSW 2007, Australia}

\begin{abstract}
Graph neural networks (GNNs) have shown great success in learning from graph-based data. The key mechanism of current GNNs is message passing, where a node's feature is updated based on the information passing from its local neighbourhood. A limitation of this mechanism is that node features become increasingly dominated by the information aggregated from the neighbourhood as we use more rounds of message passing. Consequently, as the GNN layers become deeper, adjacent node features tends to be similar, making it more difficult for GNNs to distinguish adjacent nodes, thereby, limiting the performance of GNNs. In this paper, we propose conditional local feature encoding (CLFE) to help prevent the problem of node features being dominated by the information from local neighbourhood. The idea of our method is to extract the node hidden state embedding from message passing process and concatenate it with the nodes feature from previous stage, then we utilise linear transformation to form a CLFE based on the concatenated vector. The CLFE will form the layer output to better preserve node-specific information, thus help to improve the performance of GNN models. 
To verify the feasibility of our method, we conducted extensive experiments on seven benchmark datasets for four graph domain tasks: super-pixel graph classification, node classification, link prediction, and graph regression. The experimental results consistently demonstrate that our method improves model performance across a variety of baseline GNN models for all four tasks.
\end{abstract}

\begin{keyword}
graph neural networks \sep graph representation learning \sep conditional local feature encoding
\end{keyword}

\end{frontmatter}

\section{Introduction} \label{sec:intro}

In the last few years, motivated by the success of applying deep learning techniques in Euclidian structured data, there has been an emerging interest in adapting deep learning techniques to graph-structured learning because of its great expressive power \citep{bresson2017residual, velickovic2018graph, wu2019simplifying, zhang2022ssfg, zhang2023learning}. Graph neural networks (GNNs) are a general model that utilises graph convolution for representation learning on graph-structured data. Due to their powerful performance, GNNs have become one of the dominant approaches for various graph domain tasks. Many types of practical research have been devoted to improving the performance of the GNNs in diverse areas, including social networks, knowledge graphs, chemistry molecules, and computer vision.

Unlike Euclidean-structured data like images or texts, graph-structured data have a complicated topographical structure. Each node in a graph does not have a fixed location and may have a various number of neighbourhood nodes. Therefore, convolutional neural networks (CNNs) and recurrent neural networks (RNNs), which are commonly used for processing Euclidean structured data,  are not competent for graph learning tasks. GNNs are designed to analyse graph-structured data by using graph convolution. The message passing paradigm explain the graph convolution as a process which recursively update the node's features by aggregating information from its neighbourhood nodes. GNNs are shown to be effective in a variety of graph analysis tasks, including node classification, node clustering, edge prediction, graph regression, and graph classification.

Although many approaches have been made to improve the performance of GNNs, some significant issues are still limiting their progression. 
 Due to graph convolution process, neighbourhood nodes' features are aggregated into central nodes to update the central node's representation. As the model layer become deeper, the aggregated features from the neighbourhood nodes will dominate the central node's specific representation, weakening node's original feature \citep{hamilton2020graph}. This phenomenon increases the difficulty of GNN models to learn the representation of different nodes, as the most part of one node's representation comprises the information from its neighbourhood nodes instead of node's original feature. Consequently, adjacent nodes tend to have very similar representation. \citep{li2018deeper}. 
 A commonly method for alleviating this phenomenon is to use skip connections, which can preserve the central node's information in the updated representation. Skip connections are widely used in various GNN models to preserve the node's local representation and also accelerate the training process economically \citep{xu2021optimization}. Skip connection directly preserves the locality information node representations by propagating the node feature from the previous layer to the current layer output and updating the current node feature with the consideration of the previous node information. It can be integrated with many kinds of graph convolution methods. However, since the skip connection directly propagates and combines the node feature from previous layer with the current layer's output, it fails to consider any relationship between the adjacent layers. In other words, it fails to consider the interdependency between different size of node's neighbourhood. As a result, the improvement of the skip connection is limited, and it tends to be more useful in tasks that need to predict the node based on their neighbourhood representation, e.g. node classification, but perform poorly on tasks like link prediction, which need to consider multi-layer relationships. Also, the improvement of skip connection is decreased with the increase of the model layer number. As a result, the layer number of GNN models using the skip connection method is usually limited to 5 layers to obtain economic performance \citep{hamilton2020graph}.

In this paper, we propose a conditional local feature encoding (CLFE) method to preserve the interdependency between adjacent layers by combining the previous layer output with the current layers hidden state feature. Our idea is to use concatenation and linear transformation operator to extract and combine the layer's input feature and the layer's hidden state feature, thereby obtaining the CLFE of the node features. To be specific, in each GNN layer, the nodes' hidden state feature will be concatenated with the nodes' input feature, and then the concatenated feature will be transformed through a linear transformation weight matrix to obtain a CLFE. This CLFE is then added back to the hidden state feature to form the output node representation of this layer. Our method considers the relationship between adjacent layers, which the skip connection doesn't have. Our method can be integrated into various GNN models as a plugin, and improve the model performance on multiple graph domain tasks.

We present a series of experiments in various GNN models and datasets to test our method's feasibility. We applied our method to five commonly used GNN models: graph convnets (GCN) \citep{kipf2017semi}, gated graph convnet (GatedGCN) \citep{bresson2017residual}, GraphSAGE \citep{hamilton2017inductive}, MoNet \citep{monti2017geometric}, and graph attention network (GAT) \citep{velickovic2018graph}. We tested these adapted GNN models' performance on seven benchmark datasets across four graph domain tasks, i.e., CLUSTER and PATTERN \citep{dwivedi2020benchmarking} for node classification task, CIFAR10 \citep{krizhevsky2009learning} and MNIST \citep{lecun1998gradient} for super-pixel graph classification task, TSP \citep{dwivedi2020benchmarking} and OGBL-COLLAB \citep{hu2020open} for edge classification task, and ZINC \citep{irwin2012zinc} for graph regression task. The experiment results show that our proposed method has general performance improvements in a variety of GNN models and datasets. Unlike usual skip connection method which mainly perform improvement on node classification task, our proposed method yields consistent performance improvement across four different graph domain tasks. By increasing the number of layers in the models, the experiment results also indicate that our method can achieve better performance as GNN models become deeper.

To sum up, the main contributions of this paper are listed as follows:

\begin{itemize}
  \item We propose the CLFE method that enhances the central nodes' representation in the message passing scheme. Our method concatenate the node hidden state embedding with the node embedding from previous stage, then utilise a linear transformation operator to form the CLFE. The CLFE of current layer will be added with the hidden state embedding to form the final output of the current layer. This method can enhance the central node's representation to improve the performance of GNN models.

  \item CLFE is a generic method that can be applied to graph convolutional layers in various GNN architectures as a plugin manner to improve their performance on multiple kinds of tasks. We experimentally evaluate the feasibility of our method on five commonly used GNN models across seven benchmark datasets. The experiment results across four graph domain tasks demonstrate that our method consistently improves the performance of all base GNN models.
\end{itemize}

\section{Related Work}

GNNs are novel deep-learning techniques focused on graph-structured data analytics. Through several years of work, GCNs have many variants, and different variants utilise different methods to conduct the aggregation and update processes.

\textbf{Intra-layer design:} Based on the methods of conducting graph representation learning, GNNs can be categorised as spectral methods and spatial methods.Spectral methods treat graph-structured data as signals and analyse the spectral representation of the data. \cite{bruna2014spectral} proposed the first spectral approach of GNNs. It conducts the graph convolution based on the spectrum of the graph Laplacian. \cite{kipf2017semi} proposed GCN. It introduced the first-order approximation of ChebNet, which simplified the convolution process. Besides reducing the overfitting issue of the GNNs, GCN also proposed a renormalisation method to handle the vanishing/exploding gradient and the unstable parameter problem. Spatial methods define convolution operation through the spatial relations of the nodes. \cite{gilmer2017neural} formulated a general framework of spatial methods called Message Passing Neural Networks (MPNNs). It describes the graph convolutions in a message passing perspective that the representation of one node can be passed to another node along the edge between them. Since the spatial methods treat neighbourhood node representation separately, maintaining the local invariant of the graph convolutions for different neighbourhood sizes can be challenging. As a result, the sampling method is a noticeable area in spatial GNNs. \cite{hamilton2017inductive} proposed GraphSage address a general inductive node embedding framework for large-scale graph data. It samples a fixed number of neighbourhood nodes to participate in the aggregation through a nodewise sampling method. GraphSage is also one of the first models to explore the benefit of using skip connections combined with linear interpolation in the upgrade function. 
A graph representation learning framework proposed by \cite{zhang2023learning} enables the learning and propagation of edge features to generate node embeddings instead of aggregates the node features only.
 Besides, the attention mechanism is also adapted in the area of GNN. GAT proposed by \cite{velickovic2018graph} utilises the masked self-attention tool to perform the node classification task of the graph data. RandAlign is a normalisation method proposed by \cite{zhang2024randalign}, it using random interpolation in each graph convolution layer to randomly aligns the learned embedding for each node that genertaed by the previous layer.

\textbf{Inter-layer design:} Besides the methods of conducting aggregation and update process, some of the work also focus on the inter-layer design of GNNs, in another word, the methods of conducting residual/skip connections. \cite{li2019deepgcns}, illustrated that GCNs are limited to shallow layers due to the vanishing gradient problem. They adopted the concepts of ResNet and DenseNet and proposed a new GNN model, DeepGCNs, which can be trained in very deep layers. AirGNN, proposed by \cite{liu2021graph}, found that the normal residual connection in the message passing scheme has vulnerability against abnormal node features. They introduced a interpretable and adaptive message passing scheme that enhances the network's resilience to abnormal node features on node classification task. \cite{zhang2022model} disentangled the graph convolution process into two operations: propagation and transformation, and they have found out that when a GNN is stacked with many layers, large propagation depth is the main cause of the performance degradation. Based on that finding, they proposed Adaptive Initial Residual (AIR), to tackle the performance degredation and the over-smoothing issue on node classification task.

\section{Methodology}
\subsection{Preliminary}
A graph $G=(V, E)$ can be defined by a set of nodes $V$ and a set of edges$E \subseteq V \times V$ connecting node pairs together. $N = |V|$, representing the total number of nodes in $G$. 
If an edge ${e}_{i, j} \in E$, where $i, j \in N$, it represents that ${n}_{i}$ and ${n}_{j}$ are connected together with edge ${e}_{i, j}$. A convenient way to represent nodes connectivity is by using the adjacency matrix $A \in \mathbb {R}^{N, N}$. For example, if ${A}_{i, j}$ equals to 1, it means ${e}_{i, j} \in E$, ${n}_{i}$ and ${n}_{j}$ are connected, otherwise, ${A}_{i, j}$ equals to 0. 
If one graph $G$ only contains undirected edges, then the adjacency matrix $A$ of this graph is a symmetric matrix. The degree of each node $v_{i}$ in an undirected graph $G$ can be represented as ${d}_{i}$, indicating how many edges are connecting with this node. And $D \in {\mathbb R}^{N, N}$ is a diagonal matrix representing every node's degree of graph $G$. 

Assuming a self-looped graph represented as $\tilde{G}=(V, \tilde{E})$, then the diagonal degree matrix $\tilde{D}$ of $\tilde{G}$ can be represented as $\tilde{D} = D + I$,  and the adjacency matrix of $\tilde{G}$ can be defined as $\tilde{A} = A + I$, where $I$ is the identity matrix. Then the symmetrically normalised adjacency matrix can be represented as $\tilde{A}_{sym} = \tilde{D}^{-\frac{1}{2}}\tilde{A}\tilde{D}^{-\frac{1}{2}}$. Assuming ${H}^{l}$ defines the feature representation output of the $l$-th layer, ${H}^0={X}$. Then the forward propagation of the GCN layer can be defined as follow:\begin{linenomath}\begin{equation} \label{eq: GCN}
\begin{split}
	\bm{H}^{l+1} &= \sigma(\tilde{\bm{A}}_{sym}\bm{H}^{l}\bm{W}^{l}),\ \ \ \tilde{\bm{A}}_{sym} = \tilde{\bm{D}}^{-\frac{1}{2}}\tilde{\bm{A}}\tilde{\bm{D}}^{-\frac{1}{2}},
\end{split}
\end{equation}
\end{linenomath}
where $\sigma$ is the activation function and ${W}^{l}\in\mathbb{R}^{C, f}$ is the learnable transformation matrix from the neural network in the $l$-th layer.

\ifshowfig
\begin{figure*}[h]
  \centering
  \hspace*{0cm}\includegraphics[width=0.8\textwidth]{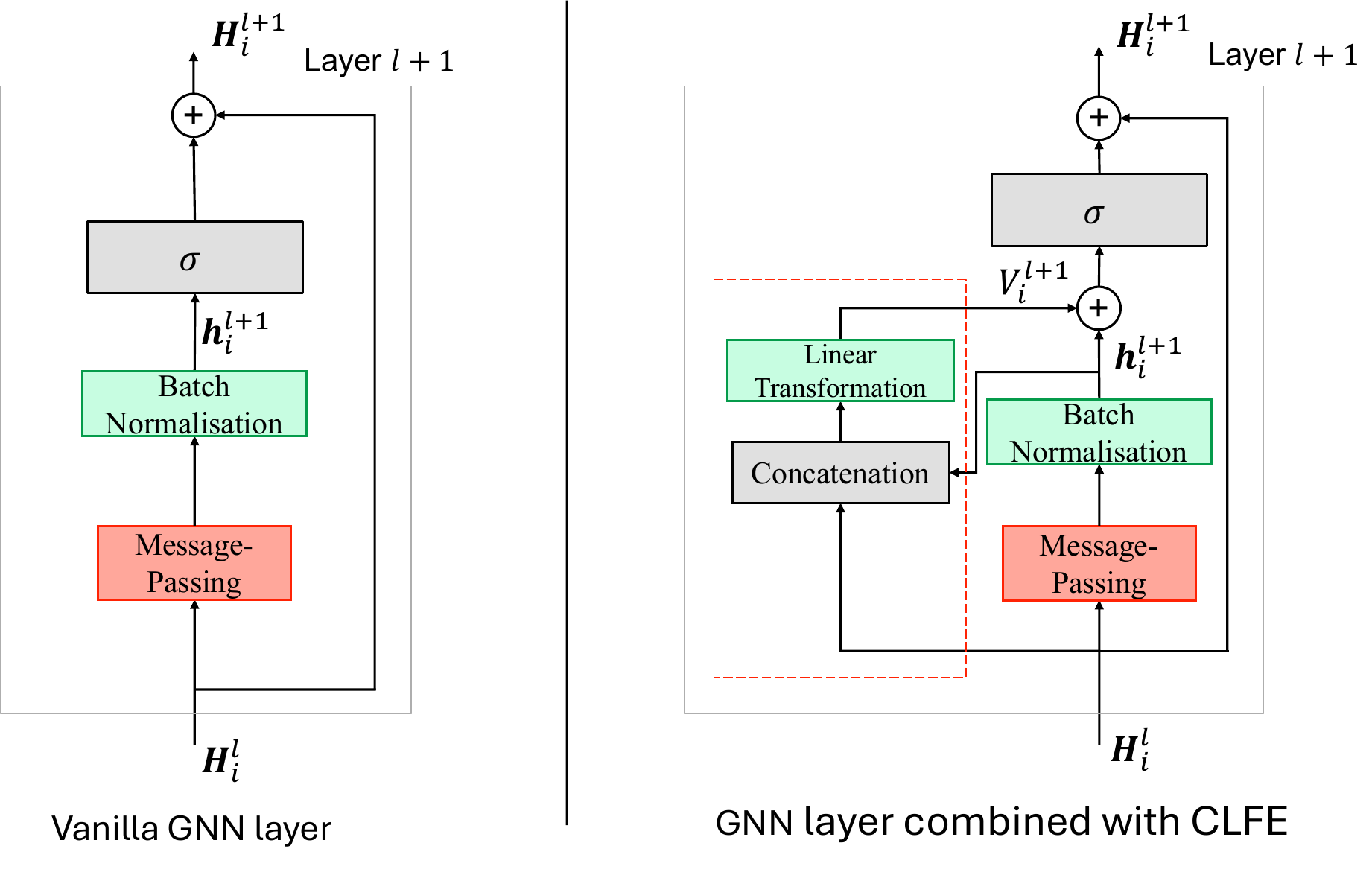}
  \caption{Proposed method combined with Layer $l+1$ from general GNN model, left hand side represents a vanilla GNN layer with skip connection, and the right hand side of the figure illustrates the GNN layer combined with our method. The red dotted block contains the CLFE plugin component in each GNN layer.}
  \label{fig:cibbectuib}
\end{figure*}
\fi

\subsection{The Proposed Conditional Local Feature Encoding}

From the perspective of the message passing framework, in each graph convolution layer, every node propagates its node features to the adjacent nodes through the edges connecting them. This process aims to let each node obtain the local feature information around them. Then, the original node representation of each node will be updated according to the local feature information from its neighbourhood. By stacking $L$ layers in the GNN models, the node features will further propagate to $L$ hops adjacent nodes, the receptive field of each node will be expanded, and the final node representation will contain $L$-hop of neighbourhood information. Based on this, Equation \ref{eq: GCN} can rewriten as:
\begin{linenomath}\begin{equation}
\begin{multlined}
	\bm{H}_{i}^{l+1} = Update \bigl(\bm{H}_{i}^{l}, Aggregate (\bm{h}_{j}^{l}, \\ \forall j \in \mathcal{N}_{(i)}) \bigr),
\end{multlined}
\end{equation}\end{linenomath}
where $\bm{H}_{i}^{l}$ represents the node feature of node $i$ in layer $l$, $\bm{h}_{j}^{l}$ represents the neighbourhood nodes' representation of node $i$ in layer $l$, $\mathcal{N}_{(i)}$ represents the neighbourhood node  set of node $i$. The aggregate function takes the node feature from previous layer as the input and aggregate the adjacent nodes' feature information. The update function utilise a learnable transformation matrix and non-linear activation function to update the node feature according to the output of aggregate function.

Although the aggregate and update process is the most commonly used message passing framework, this framework has a limitation: with more layers stacked, the neighbourhood node feature aggregating into the node's representation will become dominant in the final node representation instead of the node's original representation. This issue limits the performance of GNN models. Most of the GNN models utilise skip connection to directly preserve the local information of each message passing scheme to alleviate this issue \citep{hamilton2020graph}.
But the benefit brought by the skip connection method is limited. Most skip connection methods have limited improvement when the model layer number exceeds 5. And the improvement on different tasks is also uneven: skip connection usually performs better on tasks with homophily of the neighbourhood nodes, \emph{i.e.} node classification tasks.

To further enhance the central node's local information, we propose CLFE to encourage the reuse of information from previous layer to preserve more node-level local information during the message passing process. As Figure \ref{fig:cibbectuib} shows, the CLFE method first extracts the generated hidden state embedding of node $i$: $\bm{h}_{i}^{l+1}$, and concatenates it with the previous layer's output $\bm{H}_{i}^{l}$ to avoid the entanglement of the information during the message passing process. Then the concatenation matrix will be transformed by the linear transformation operator with learnable weight matrix $\bm{W}^{l}$ and bias $b$, the output of the linear transformation operator: the nodes' CLFE $\bm{V}_{i}^{l}$ will be added with the hidden state embedding $\bm{h}_{i}^{l+1}$ to form the final output of layer $l+1$. The local feature encoding $\bm{V}_{i}^{l}$ can be described as follows:

\begin{linenomath}\begin{equation} \label{eq:proposedmethod1}
\begin{aligned}
	\bm{V}_{i}^{l} &= Concat(\bm{H}_{i}^{l}, \bm{h}_{i}^{l+1}) \bm{W}^{l} + b, \\
\end{aligned}
\end{equation}\end{linenomath}
where $Concat(\cdot)$ denotes the concatenation operator, $\bm{W}^{l}$ and $b$ are learnable weight matrix and bias of the linear transformation, respectively. To generate the final output feature of current layer in the GNN model, $\bm{V}_{i}^{L}$ will be added with the generated hidden state node embedding $\bm{h}_{i}^{l+1}$ to form the final node feature:

\begin{linenomath}
\begin{equation} \label{eq:proposedmethod2}
	\bm{H}_{i}^{l+1} = \sigma (\bm{V}_{i}^{l}+\bm{h}_{i}^{l+1}) + \bm{H}_{i}^{l},
\end{equation}
\end{linenomath}
where $\bm{H}_{i}^{l}$ is the node representation from last layer propogated by the skip connection. In this way, our method will extract the node's local feature encoding in each convolution layer of the GNN models and generate the final node feature by adding the local feature encoding back to the update process.
Our method can be combined with various graph architectures as a plugin. It conducts conditional encoding in each convolution layer and captures the most useful representation based on the node's local neighbourhood. 

We also modified the prediction layer based on four graph domain tasks to generate task-dependent outputs. For super-pixel graph classification task, we utilise a readout function to average over all node features from the last GNN layer to form a graph-level representation vector. The vector is then passed to a 2-layer MLP to generate the logits for each graph class. During training, we minimise the difference between predicted outputs and ground truth labels using cross-entropy loss.
For graph regression task, similarly with super-pixel graph classification task, we generate the graph-level representation vector using a readout function from node features from the final GNN layer. The graph representation is then passed to a 2-layer MLP to generate the prediction score. The L1-loss between the predicted score and the groundtruth score is used for optimisation during the training.
For node classification task, feature vectors of each node are independently passed to a 2-layer MLP to generate the prediction for each class. We then minimise the cross-entropy loss between the predictions and the groundtruth scores to improve the performance.
For edge prediction task, we concatenate the features of two adjacent nodes from the final GNN layer to form the edge feature vector, then we pass the edge feature vector to a 2-layer MLP to compute the logit for each class. Cross-entropy loss is utilised to minimise the loss between the predicted results and the groundtruth labels.

\section{Experiments}

\subsection{Experimental Setup}

\begin{figure*}[h]
	\centering
	\captionsetup{justification=centering}
	\begin{subfigure}{0.3\textwidth}
		\centering
		\includegraphics[width=1\linewidth]{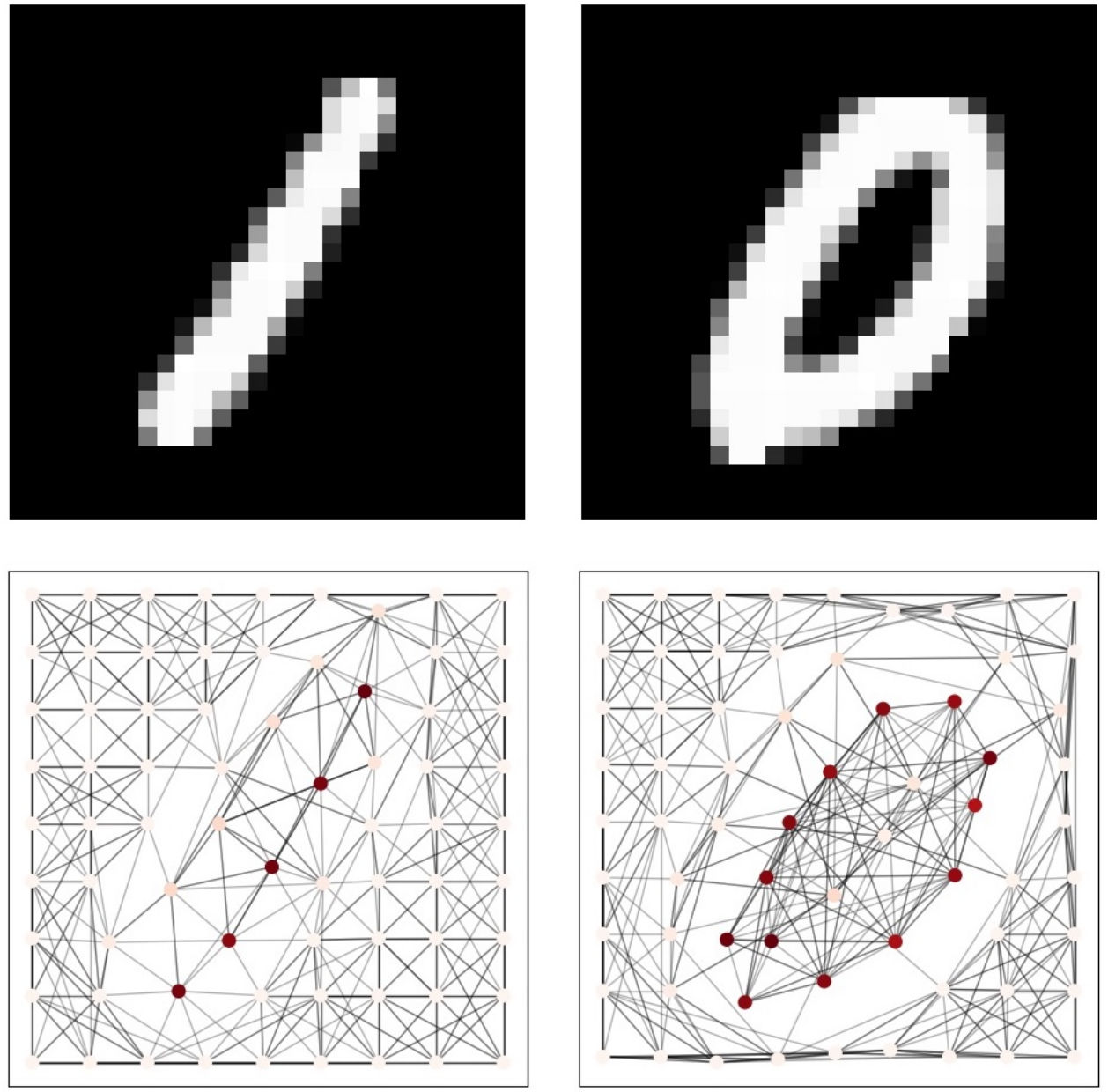}
		\caption{MNIST with label 1 and 0}
		\label{fig: samples-mnist}
	\end{subfigure}
	\hspace{4em}
	\begin{subfigure}{0.3\textwidth}
		\centering
		\includegraphics[width=1\linewidth]{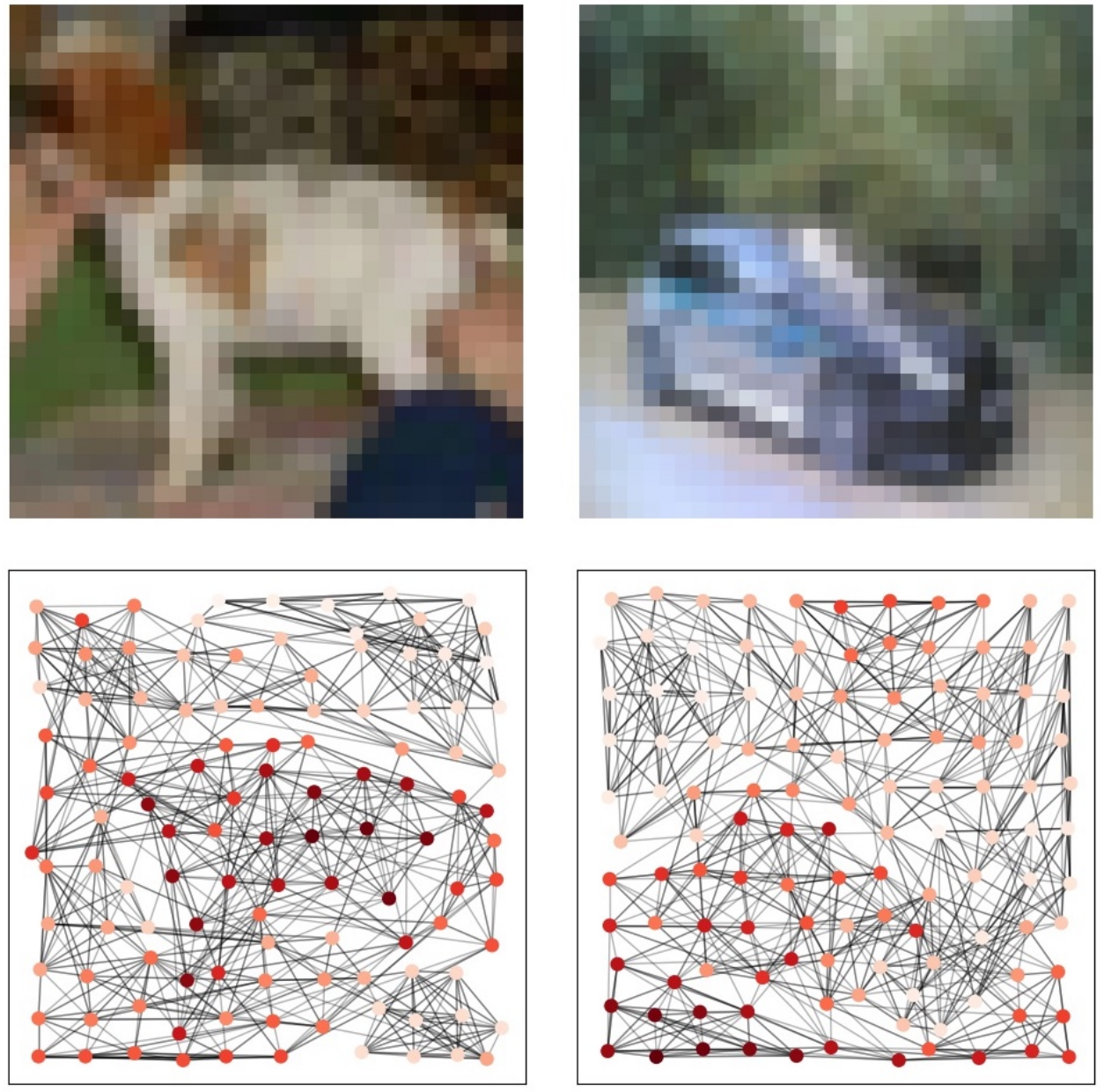}
		\caption{CIFAR10 with label dog and car}
		\label{fig: samples-cifar}
	\end{subfigure}
	
\caption{Samples of images (first row) and their corresponding superpixel graphs (second row) from MNIST and CIFAR10. The feature of a node consists of the superpixel's coordinate and intensity.}
\label{fig: samples}
\end{figure*}

\textbf{Datasets.} To verify the effectiveness of our proposed method, we utilise seven benchmark datasets that cover four graph-based tasks: node classification, super-pixel graph classification, link prediction, and graph regression. Detailed statistics of these seven datasets are illustrated in Table \ref{table:statics}. 
\begin{itemize}
	\item \textbf{PATTERN and CLUSTER} are generated with the Stochastic Block Model, which is proposed by \cite{abbe2017community}, a commonly utilised model to analyse the social network in the real world. These two datasets are used in our experiment to test the model performance on the node classification task. Thereinto, the PATTERN dataset conducts the node-level tasks of graph pattern recognition, and the CLUSTER dataset conducts the semi-supervised graph clustering.
	\item \textbf{MNIST and CIFAR-10} are commonly used in computer vision for the image classification task. In this paper, these two datasets will be preprocessed as super-pixel datasets, examples are shown in Figure \ref{fig: samples}. The super-pixels of one image can be extracted by using the SLIC method \citep{achanta2012slic}, and they represent a fraction of an image that has homogeneous intensity.
	\item \textbf{Travelling Salesman Problem (TSP)} conducts the binary edge classification task following the work of \cite{dwivedi2020benchmarking}. The dataset evaluates the performance of the GNN model to find an optimal sequence of nodes with a minimal edge weight.
	\item \textbf{OGBL-COLLAB} contains one large collaboration network that models the collaborative relationships between scientists. This graph has labelled the collaboration year into the edges' feature. In our experiment, we conduct the edge classification task by using the collaborations until 2017 as the training set, collaborations until 2018 as the validation set and those in 2019 as the test set \citep{hu2020open}.
	\item \textbf{ZINC} is a chemistry dataset about molecular graphs \citep{irwin2012zinc}. Following the work of \cite{dwivedi2020benchmarking}, we use a subset (12K) of this dataset to test the graph regression ability of our method. Each subgraph of ZINC represents one molecular, the node features represent the atomic class, and the edges represent the type of chemical bonds.
\end{itemize}

\begin{table*}[tbp]
\caption{Statistics of the seven benchmark datasets for our experiments.}

\centering  
\begin{adjustbox}{width=\linewidth, center}
\begin{tabular}{ l | c | cccccc }
\toprule[1pt]
Dataset  &Graphs & Nodes/graph &\#Training &\#Validation &\#Test &\#Categories & Task\\

\midrule[.6pt]

PATTERN & 14K &44-188 &10,000 &2,000 &2,000 & 2 &\multirow{2}{*}{ Node Classification}\\
CLUSTER & 12K &41-190 &10,000 &1,000 &1,000 & 6 & \\

\midrule[.5pt]
MNIST   &70K &40-75  &55,000 &5000 &10,000 & 10 &\multirow{2}{*}{ Super-pixel Graph Classification}\\

CIFAR10 &60K &85-150 &45,000 &5000 &10,000 & 10 &\\

\midrule[.5pt]

TSP    & 12K & 50-500 & 10,000 & 1,000 & 1,000 & -- &\multirow{2}{*}{Edge Classification}\\

OGBL-COLLAB & 1 & 235,868 & \\

\midrule[.5pt]

ZINC &12K &9-37 &10,000 &1,000 &1,000 & -- & Graph regression\\

\bottomrule[1pt]

\end{tabular}
\end{adjustbox}
\label{table:statics}
\end{table*}

\textbf{Evaluation Metrics.} Different tasks have different corresponding evaluation metrics. Four evaluation metrics utilised in this experiment are listed below:
\begin{itemize}
	\item \textbf{Accuracy.} We use the weighted average accuracy to evaluate the performance for node classification tasks conducted in the PATTERN and CLUSTER datasets and graph classification tasks conducted in the MNIST and CIFAR10 datasets.
	\item \textbf{F1 Score.} For the edge classification task conducted in the TSP dataset, we use the F1 score for the positive class to evaluate the performance since the positive labelled edges only take a small part of the overall edges inside one graph.
	\item \textbf{Hits@K.} For the edge classification conducted in the OGBL-COLLAB dataset, we use Hits@K. This metric ranks each positive prediction against 100000 randomly sampled negative collaborations and calculates the ratio of positive edges that are ranked at K-th place or above.
	\item \textbf{MAE.} The mean absolute error is used for evaluating the graph regression task of the ZINC dataset. It measures each molecular graph's predicted and ground truth-constrained solubility.
\end{itemize}

\textbf{Implementation Details.}
The experiment implementation is based on the work of \cite{dwivedi2020benchmarking} using PyTorch with the DGL library \cite{wang2019deep}.
During the training process, the learning rate will be linearly decayed until the validation loss no longer decreases in a fixed number of epochs (usually 5 or 10, depending on different tasks). The Adam algorithm \citep{kingma2014adam} is utilised for training the models as optimizers.
The initial learning rate, decay factor, and training patience may vary depending on different experiments and different GNN models. The training process will be terminated when the learning rate reaches the minimum value of $10^{-6}$. 
The experiments are conducted on a Quadro RTX 6000 GPU with 24 GB memory.
We validate our models with different numbers of graph convolutional layers, \emph{e.g.,} 4 and 16 in the node classification task.
For each evaluation, we conduct experiments four times using four different random seeds to test the robustness of the models, \emph{e.g.} 9, 23, 41, and 42. We report the mean value and the standard deviation between the four runs.
\subsection{Experimental Results}
 The corresponding quantitive experimental results of the five GNN models with the seven datasets are shown in Table \ref{table:MNIST} - \ref{table:ZINC}. The best test result across all five models in one dataset without using our proposed method is highlighted with violet colour. And the best test result across all five models using our proposed method in one dataset is highlighted in bold blue. After adding the proposed method, the performance changes are shown in the parenthesis after the test result, with arrows indicating an increase or decrease. In the performance changes of the proposed method, the improvements which are larger than 3 per cent are highlighted with cerulean colour.

\begin{table*}[tbp]
\caption{Performance of our model for graph classification on MNIST.}
\centering 
\begin{adjustbox}{width=\linewidth, center}

\begin{tabular}{ l cccc cccc }
\toprule[1pt]

\multirow{2}{*}{ Method} &\multirow{2}{*}{$L$}    & \multicolumn{2}{c}{Test Accuracy}  & \multicolumn{2}{c}{Training Accuracy} \\
\cmidrule(lr){3-4} \cmidrule(lr){5-6}
 & &w/o CLFE &+ CLFE &w/o CLFE &+ CLFE\\
\midrule[.6pt]
GCN \citep{kipf2017semi} &4 &90.705$\pm$0.218 &96.943$\pm$0.129 ({\color{cerulean}6.238$\uparrow$}) &97.196$\pm$0.223 &99.999$\pm$0.001 (2.803$\uparrow$)\\
GraphSAGE  \citep{hamilton2017inductive}   &4 &97.312$\pm$0.097 &97.725$\pm$0.175 (0.413$\uparrow$) &100.00$\pm$0.000 &100.00$\pm$0.000 (0.000$\uparrow$)   \\
\midrule[.6pt]
MoNet \citep{monti2017geometric} &4 &90.805$\pm$0.032 &96.168$\pm$0.230 ({\color{cerulean}5.363$\uparrow$}) &96.609$\pm$0.440&99.988$\pm$0.016 (3.379$\uparrow$)\\
GAT       \citep{velickovic2018graph}  &4 &95.535$\pm$0.205 &97.115$\pm$0.117 (1.580$\uparrow$) &99.994$\pm$0.008 &100.00$\pm$0.000 (0.006$\uparrow$)\\
GatedGCN \citep{bresson2017residual} &4 &{\color{violet}97.340$\pm$0.143} &{\color{blue}\textbf{97.912$\pm$0.078}} (0.572$\uparrow$) &100.00$\pm$0.000 &100.00$\pm$0.000 (0.000$\uparrow$)\\

\bottomrule[1pt]

\end{tabular}

\end{adjustbox}
\label{table:MNIST}
\end{table*}

\begin{table*}[htbp]
\caption{Performance of our model for graph classification on CIFAR10.}

\centering  
\begin{adjustbox}{width=\linewidth, center}

\begin{tabular}{ l cccc cccc }
\toprule[1pt]

\multirow{2}{*}{ Method} &\multirow{2}{*}{$L$}    & \multicolumn{2}{c}{Test Accuracy}  & \multicolumn{2}{c}{Training Accuracy} \\
\cmidrule(lr){3-4} \cmidrule(lr){5-6}
 & &w/o CLFE &+ CLFE &w/o CLFE &+ CLFE\\
\midrule[.6pt]
GCN \citep{kipf2017semi} &4 &55.710$\pm$0.381 &63.208$\pm$0.401 ({\color{cerulean}7.498$\uparrow$}) &69.523$\pm$1.948 &88.040$\pm$2.748 (18.517$\uparrow$)\\
GraphSAGE \citep{hamilton2017inductive}   &4 &65.767$\pm$0.308 &66.633$\pm$0.284 (0.866$\uparrow$) &99.719$\pm$0.062 &99.895$\pm$0.096  (0.176$\uparrow$)\\
\midrule[.6pt]
MoNet \citep{monti2017geometric} &4 &54.655$\pm$0.518 &62.072$\pm$0.157 ({\color{cerulean}7.417$\uparrow$}) &65.911$\pm$2.515 &81.173$\pm$0.967 (15.262$\uparrow$)\\
GAT       \citep{velickovic2018graph}  &4 &64.223$\pm$0.455  &66.706$\pm$0.377 (2.473$\uparrow$) &89.114$\pm$0.499 &99.896$\pm$0.081 (10.752$\uparrow$)\\
GatedGCN \citep{bresson2017residual} &4 &{\color{violet}67.312$\pm$0.311} &{\color{blue}\textbf{68.325$\pm$0.308}} (1.013$\uparrow$) &94.553$\pm$1.018 &99.998$\pm$0.001 (5.445$\uparrow$) \\

\bottomrule[1pt]

\end{tabular}
\end{adjustbox}

\label{table:CIFAR10}
\end{table*}

\begin{table*}[tbp]
\caption{Performance of our model for node classification on PATTERN.}

\centering  
\begin{adjustbox}{width=\linewidth, center}

\begin{tabular}{ l cccc cccc }
\toprule[1pt]

\multirow{2}{*}{ Method} &\multirow{2}{*}{$L$}  & \multicolumn{2}{c}{Test Accuracy}  & \multicolumn{2}{c}{Training Accuracy} \\
\cmidrule(lr){3-4} \cmidrule(lr){5-6}
 & &w/o CLFE &+ CLFE &w/o CLFE &+ CLFE\\
\midrule[.6pt]
GCN \citep{kipf2017semi} &4  &63.880$\pm$0.074 &68.836$\pm$0.119 ({\color{cerulean}4.956$\uparrow$}) &65.126$\pm$0.135 &73.103$\pm$0.356 (7.977$\uparrow$)\\
                        &16 &71.892$\pm$0.334 &79.676$\pm$1.617 ({\color{cerulean}7.784$\uparrow$}) &78.409$\pm$1.592 &89.694$\pm$2.798 (11.285$\uparrow$)\\
GraphSAGE  \citep{hamilton2017inductive}   &4 &50.516$\pm$0.001 &50.516$\pm$0.003 (0.000$\uparrow$) &50.473$\pm$0.014  &50.479$\pm$0.009 (0.006$\uparrow$)\\
                                          &16 &50.492$\pm$0.001 &50.493$\pm$0.002 (0.001$\uparrow$) &50.487$\pm$0.005 &50.483$\pm$0.010 (0.004$\downarrow$)       \\

\midrule[.6pt]
MoNet \citep{monti2017geometric} &4 &85.482$\pm$0.037 &85.705$\pm$0.027 (0.223$\uparrow$) &85.569$\pm$0.044 &85.720$\pm$0.048 (0.151$\uparrow$)\\
                                &16 &{\color{violet}85.582$\pm$0.038} &85.793$\pm$0.035 (0.211$\uparrow$) &85.720$\pm$0.068 &86.440$\pm$0.130 (0.720$\uparrow$)\\
GAT       \citep{velickovic2018graph}  &4 &75.824$\pm$1.823 &79.564$\pm$0.556 ({\color{cerulean}3.740$\uparrow$}) &77.883$\pm$1.632 &85.128$\pm$0.435 (7.245$\uparrow$) \\
                                      &16 &78.271$\pm$0.186 &80.079$\pm$0.107 (1.808$\uparrow$) &90.212$\pm$0.476 &99.115$\pm$0.163 (8.903$\uparrow$) \\
GatedGCN \citep{bresson2017residual} &4 &84.480$\pm$0.122 &85.473$\pm$0.029 (0.993$\uparrow$) &84.474$\pm$0.155 &85.390$\pm$0.049 (0.916$\uparrow$) \\
                                    &16 &85.568$\pm$0.088 &{\color{blue}\textbf{86.383$\pm$0.095}} (0.815$\uparrow$) &86.007$\pm$0.123 &87.449$\pm$0.357 (1.442$\uparrow$) \\

\bottomrule[1pt]

\end{tabular}
\end{adjustbox}

\label{table:PATTERN}
\end{table*}

\begin{table*}[tbp]
\caption{Performance of our model for node classification on CLUSTER.}

\centering  
\begin{adjustbox}{width=\linewidth, center}

\begin{tabular}{ l cccc cccc }
\toprule[1pt]

\multirow{2}{*}{Method} &\multirow{2}{*}{$L$}    & \multicolumn{2}{c}{Test Accuracy}  & \multicolumn{2}{c}{Training Accuracy} \\
\cmidrule(lr){3-4} \cmidrule(lr){5-6}
 & &w/o CLFE &+ CLFE  &w/o CLFE &+ CLFE\\
\midrule[.6pt]
GCN \citep{kipf2017semi} &4 & 53.445$\pm$2.029 &59.095$\pm$0.038 ({\color{cerulean}5.65$\uparrow$}) &54.041$\pm$2.197 & 61.955$\pm$0.166 (7.914$\uparrow$)\\
    &16 &68.498$\pm$0.976 &74.330$\pm$0.106 ({\color{cerulean}5.832$\uparrow$}) &71.729$\pm$2.212 & 84.382$\pm$0.318 (12.635$\uparrow$)\\
GraphSAGE  \citep{hamilton2017inductive}   &4 &50.454$\pm$0.145 &50.709$\pm$0.187 (0.255$\uparrow$) &54.374$\pm$0.203 &54.756$\pm$0.222 (0.382$\uparrow$)\\
                                          &16 &63.844$\pm$0.110 &67.200$\pm$0.455 (3.356$\uparrow$) &86.710$\pm$0.167 &85.110$\pm$0.501 (1.600$\downarrow$)    \\

\midrule[.6pt]
MoNet \citep{monti2017geometric} &4 &58.064$\pm$0.131 &65.588$\pm$0.095 ({\color{cerulean}7.524$\uparrow$}) &58.454$\pm$0.183 &67.433$\pm$0.125 (8.979$\uparrow$)\\
                                &16 & 66.407$\pm$0.540 &74.571$\pm$0.576 ({\color{cerulean}8.164$\uparrow$)} &67.727$\pm$0.649 &76.981$\pm$1.405 (9.254$\uparrow$) \\
GAT       \citep{velickovic2018graph}  &4 & 57.732$\pm$0.323 &59.401$\pm$0.095 (1.669$\uparrow$) &58.331$\pm$0.342 &62.597$\pm$0.650 (4.226$\uparrow$)\\
                                      &16 &70.587$\pm$0.447 &72.307$\pm$0.482 (1.720$\uparrow$) &76.074$\pm$1.362 &98.938$\pm$0.664 (22.864$\uparrow$)  \\

GatedGCN \citep{bresson2017residual} &4 &60.404$\pm$0.419 &63.157$\pm$0.542 (2.753$\uparrow$) &61.618$\pm$0.536 &63.864$\pm$1.094 (2.246$\uparrow$) \\
                                    &16 &{\color{violet}73.840$\pm$0.326} &{\color{blue}\textbf{75.807$\pm$0.195}} (1.967$\uparrow$) &87.880$\pm$0.908 &88.026$\pm$0.331 (0.146$\uparrow$) \\

\bottomrule[1pt]

\end{tabular}
\end{adjustbox}
\label{table:CLUSTER}
\end{table*}

\begin{table*}[tbp]
\caption{Performance of our model  for link prediction on TSP.}

\centering  
\begin{adjustbox}{width=\linewidth, center}
\begin{tabular}{ l cccc cccc }
\toprule[1pt]

\multirow{2}{*}{ Method} &\multirow{2}{*}{$L$}    & \multicolumn{2}{c}{Test F1}  & \multicolumn{2}{c}{Training F1} \\
\cmidrule(lr){3-4} \cmidrule(lr){5-6}
 & &w/o CLFE &+ CLFE  &w/o CLFE &+ CLFE\\
\midrule[.6pt]

GCN \citep{kipf2017semi} &4 &0.630$\pm$0.001 &0.683$\pm$0.006 (0.053$\uparrow$) &0.631$\pm$0.001 &0.671$\pm$0.665 (0.040$\uparrow$) \\
GraphSAGE  \citep{hamilton2017inductive}   &4 &0.665$\pm$0.003 &0.713$\pm$0.019 (0.048$\uparrow$) &0.669$\pm$0.003 &0.705$\pm$0.024 (0.036$\uparrow$)\\

MoNet \citep{monti2017geometric} &4 &0.641$\pm$0.002 &0.667$\pm$0.004 (0.026$\uparrow$) &0.643$\pm$0.002 &0.669$\pm$0.004 (0.026$\uparrow$)\\
GAT       \citep{velickovic2018graph}  &4 &0.671$\pm$0.002 &0.705$\pm$0.002 (0.034$\uparrow$) &0.673$\pm$0.002 &0.709$\pm$0.003 (0.036$\uparrow$)\\
GatedGCN \citep{bresson2017residual} &4 & {\color{violet}0.791$\pm$0.003} &{\color{blue}\textbf{0.817$\pm$0.003}} (0.026$\uparrow$) &0.793$\pm$0.003 &0.817$\pm$0.003 (0.024$\uparrow$)\\

\bottomrule[1pt]

\end{tabular}
\end{adjustbox}
\label{table:TSP}
\end{table*}

\begin{table*}[tbp]
\caption{Performance of our model for link prediction on OGBL-COLLAB.}

\centering  
\begin{adjustbox}{width=\linewidth, center}

\begin{tabular}{ l cccc cccc }
\toprule[1pt]

\multirow{2}{*}{ Method} &\multirow{2}{*}{$L$}   & \multicolumn{2}{c}{Test Hits@50}  & \multicolumn{2}{c}{Training Hits@50} \\
\cmidrule(lr){3-4} \cmidrule(lr){5-6}
 & &w/o CLFE &+ CLFE  &w/o CLFE &+ CLFE\\
\midrule[.6pt]

GCN \citep{kipf2017semi} &3 &50.422$\pm$1.131 &51.779$\pm$0.965 (1.357$\uparrow$) &92.112$\pm$0.991 &91.533$\pm$0.604 (0.579$\downarrow$)\\
GraphSAGE  \citep{hamilton2017inductive}   &3 &51.618$\pm$0.690 &52.593$\pm$0.521 (0.974$\uparrow$) &99.949$\pm$0.052 &99.945$\pm$0.020 (0.004$\downarrow$) \\

MoNet \citep{monti2017geometric} &3 &36.144$\pm$2.191 &53.825$\pm$1.156 ({\color{cerulean}17.681$\uparrow$}) &61.156$\pm$3.973 &93.405$\pm$0.060 (32.249$\uparrow$)\\
GAT       \citep{velickovic2018graph}  &3 &51.501$\pm$0.962 &54.110$\pm$0.396 (2.609$\uparrow$) &97.851$\pm$1.114 & 99.613$\pm$0.074 (1.762$\uparrow$) \\
GatedGCN \citep{bresson2017residual} &3 &52.635$\pm$1.168 & {\color{blue}\textbf{54.153$\pm$0.507}} (1.518$\uparrow$) &96.103$\pm$1.876 &99.833$\pm$0.094 (3.730$\uparrow$) \\

\bottomrule[1pt]

\end{tabular}
\end{adjustbox}

\label{table:COLLAB}
\end{table*}

\begin{table*}[tbp]
\caption{Performance of our model for graph regression on ZINC.}

\centering  
\begin{adjustbox}{width=\linewidth, center}

\begin{tabular}{ l cccc cccc }
\toprule[1pt]

\multirow{2}{*}{ Method} &\multirow{2}{*}{$L$}    & \multicolumn{2}{c}{Test MAE ($\downarrow$)}  & \multicolumn{2}{c}{Training MAE ($\downarrow$)} \\
\cmidrule(lr){3-4} \cmidrule(lr){5-6}
 & &w/o CLFE &+ CLFE  &w/o CLFE &+ CLFE\\
\midrule[.6pt]
GCN \citep{kipf2017semi} &4 &0.459$\pm$0.006 &0.401$\pm$0.002 (0.058$\downarrow$) &0.343$\pm$0.011 &0.197$\pm$0.011 (0.146$\downarrow$)\\
                        &16 &0.367$\pm$0.011 &0.310$\pm$0.010 (0.048$\downarrow$) &0.128$\pm$0.019 &0.062$\pm$0.010 (0.066$\downarrow$)\\

GraphSAGE  \citep{hamilton2017inductive}   &4 &0.468$\pm$0.003 &0.459$\pm$0.005 (0.009$\downarrow$) &0.251$\pm$0.004 &0.229$\pm$0.017 (0.023$\downarrow$)\\
                                          &16 &0.398$\pm$0.002 &0.373$\pm$0.005 (0.025$\downarrow$) &0.081$\pm$0.009 &0.094$\pm$0.004 (0.013$\uparrow$)\\

MoNet \citep{monti2017geometric} &4 &0.397$\pm$0.010 &0.315$\pm$0.007 (0.082$\downarrow$) &0.318$\pm$0.016  &0.190$\pm$0.021 (0.128$\downarrow$)\\
                                &16 &{\color{violet}0.292$\pm$0.006} &{\color{blue}\textbf{0.249$\pm$0.003}} (0.043$\downarrow$) &0.093$\pm$0.014 &0.047$\pm$0.010 (0.046$\downarrow$)\\
GAT       \citep{velickovic2018graph}  &4 &0.475$\pm$0.007 &0.410$\pm$0.010 (0.065$\downarrow$) &0.317$\pm$0.006 &0.146$\pm$0.041 (0.171$\downarrow$)\\
                                      &16 &0.384$\pm$0.007 &0.295$\pm$0.012 (0.089$\downarrow$) &0.067$\pm$0.004 &0.029$\pm$0.004 (0.038$\downarrow$)\\
GatedGCN \citep{bresson2017residual} &4 &0.435$\pm$0.011  &0.395$\pm$0.012 (0.040$\downarrow$) &0.287$\pm$0.014 &0.120$\pm$0.009  (0.167$\downarrow$)\\

\bottomrule[1pt]

\end{tabular}
\end{adjustbox}
\label{table:ZINC}
\end{table*}

The super-pixel graph classification task results with datasets MNIST and CIFAR10 are reported in Table \ref{table:MNIST} and \ref{table:CIFAR10}. As the results illustrated, the proposed method yields consistent performance improvement for five GNN models on both of the datasets. Thereinto, GCN and MoNet have gained significant improvement compared with the other three models on both MNIST and CIFAR10 datasets in terms of both test accuracy and training accuracy. On MNIST, our method improves GCN with a 6.238\% performance gain in test accuracy and MoNet with a 5.363\% performance gain in test accuracy. On CIFAR10, GCN and MoNet still exhibit significant performance improvement with the aid of our method. Specifically, GCN with CLFE has achieved a 7.498\% improvement in test accyracy and 18.517\% improvement in training accuracy, MoNet with CLFE has achieved a 7.417\% improvement in test accuracy and 15.262\% improvement in training accuracy. With the same layer number, our method improves the performance on both test accuracy and training accuracy, which exhibits our method's ability to address the underfitting problem of the model on this task.
 
 The node classification task results with datasets PATTERN and CLUSTER are reported in Table \ref{table:PATTERN} and \ref{table:CLUSTER}. In this task, we also tested our method's feasibility for enabling the model to have more layers. And it can be seen that our method can still improve the overall model's performance after the layer number is increased to 16. On PATTERN dataset, GCN with CLFE has a significant improvement in both test accuracy and training accuracy, yielding 4.956\% and 7.977\% gians. And with the layer number increased to 16, the improvement grows even further, yielding a 7.784\% improvement in test accuracy and 11.285\% in training accuracy. GAT also have a considerable improvement combined with our method, yielding a 3.740\% and 7.245\% performance gain on test accuracy and training accuracy respectively, and the improvement continues to exist when the layer number increases. The best test accuracy is achieved by GatedGCN with CLFE, yielding 86.383. On CLUSTER, our method considerably improved the performance of all five GNN models. Thereinto, GCN with 4 layers have a 5.65\% improvement in test accuracy and 7.914\% improvement in training accuracy. In GraphSAGE with 16 layers, our method increased the test accuracy by 3.356\% without increasing the training accuracy. This demonstrated that our method addressed the overfitting problem in this model on CLUSTER dataset. Other considerable improvements are made in GCN with 16 layers, and MoNet with 16 layers, yields 5.832\%, and 8.164\% performance gain in test accuracy respectively. By increased the model layer number from 4 to 16, it proves that our method enables the GNN models to have more layers without hurting the performance.
 
 The link prediction task results with datasets TSP and OGBL-COLLAB are reported in Table \ref{table:TSP} and \ref{table:COLLAB}. Our method has achieved a general improvement over all of the GNN models. Thereinto, GatedGCN, combined with our method, achieved a 0.817 F1 score on TSP dataset. On OGBL-COLLAB dataset, our method has achieved 17.681 test Hits@50 gain and 32.249 training Hits@50 gain on MoNet with 3 layer, which is the most significant improvement in our experiment. On OGBL-COLLAB dataset, our method addresses the overfitting problem in GCN and GraphSAGE, and also under fitting problem in GAT and GatedGCN.
 
 The final task is the graph regression, and the results are shown in Table \ref{table:ZINC}. We perform our experiment on the ZINC dataset and use the mean absolute error to evaluate the models' graph regression performance. Our method has generally reduced the MAE of all five GNN models. We have significantly improved the performance of GCN, MoNet, GAT, and GatedGCN. The best test MAE yields 0.249 on 16-layer MoNet with CLFE. Especially when the model goes deeper, the performance improvements persist. 
  
The experiment results prove the suitability of the proposed connection method on various types of popular GNN models. Across different tasks, our method has more considerable improvement on GCN and MoNet architecture, but the improvement on GraphSAGE is relatively limited. Our method has shown improvements not only in the node classification task but also in the other three tasks. In most of cases, our method improves the test performance of the model by adressing the underfitting and overfitting problem.

\section{CONCLUSION}

In this paper, we introduced the CLFE method to alleviate the dominance of neighbourhood node representation in the central node by further enhancing the central node's representation throughout the graph learning. We combined our method with five commonly used GNN models and experimentally validated our method on seven benchmark datasets across four graph learning tasks, including spuer-pixel graph classification, node classification, edge classification, and graph regression. The experimental results demonstrated that the overall performances across seven datasets have improved by adding our approach to each graph convolutional layer of the GNN model. After the experiments, we have found that our method helps to reduce the underfitting problem and excels in the node classification task. Our method can reduce the oversmoothing issue caused by the dominance of the neighbourhood information inside each node representation and enable the model to stack more layers. 

\section*{Acknowledgement}
We would like to thank the anonymous reviewers for reviewing this manuscript.


\begin{thebibliography}{26}
\expandafter\ifx\csname natexlab\endcsname\relax\def\natexlab#1{#1}\fi
\providecommand{\url}[1]{\texttt{#1}}
\providecommand{\href}[2]{#2}
\providecommand{\path}[1]{#1}
\providecommand{\DOIprefix}{doi:}
\providecommand{\ArXivprefix}{arXiv:}
\providecommand{\URLprefix}{URL: }
\providecommand{\Pubmedprefix}{pmid:}
\providecommand{\doi}[1]{\href{http://dx.doi.org/#1}{\path{#1}}}
\providecommand{\Pubmed}[1]{\href{pmid:#1}{\path{#1}}}
\providecommand{\bibinfo}[2]{#2}
\ifx\xfnm\relax \def\xfnm[#1]{\unskip,\space#1}\fi
\bibitem[{Abbe(2017)}]{abbe2017community}
\bibinfo{author}{Abbe, E.} (\bibinfo{year}{2017}).
\newblock \bibinfo{title}{Community detection and stochastic block models: recent developments}.
\newblock {\it \bibinfo{journal}{The Journal of Machine Learning Research}\/},  {\it \bibinfo{volume}{18}\/}, \bibinfo{pages}{6446--6531}.
\bibitem[{Achanta et~al.(2012)Achanta, Shaji, Smith, Lucchi, Fua \& S{\"u}sstrunk}]{achanta2012slic}
\bibinfo{author}{Achanta, R.}, \bibinfo{author}{Shaji, A.}, \bibinfo{author}{Smith, K.}, \bibinfo{author}{Lucchi, A.}, \bibinfo{author}{Fua, P.}, \& \bibinfo{author}{S{\"u}sstrunk, S.} (\bibinfo{year}{2012}).
\newblock \bibinfo{title}{Slic superpixels compared to state-of-the-art superpixel methods}.
\newblock {\it \bibinfo{journal}{IEEE transactions on pattern analysis and machine intelligence}\/},  {\it \bibinfo{volume}{34}\/}, \bibinfo{pages}{2274--2282}.
\bibitem[{Bresson \& Laurent(2017)}]{bresson2017residual}
\bibinfo{author}{Bresson, X.}, \& \bibinfo{author}{Laurent, T.} (\bibinfo{year}{2017}).
\newblock \bibinfo{title}{Residual gated graph convnets}.
\newblock {\it \bibinfo{journal}{arXiv preprint arXiv:1711.07553}\/}, .
\bibitem[{Bruna et~al.(2014)Bruna, Zaremba, Szlam \& Lecun}]{bruna2014spectral}
\bibinfo{author}{Bruna, J.}, \bibinfo{author}{Zaremba, W.}, \bibinfo{author}{Szlam, A.}, \& \bibinfo{author}{Lecun, Y.} (\bibinfo{year}{2014}).
\newblock \bibinfo{title}{Spectral networks and locally connected networks on graphs}.
\newblock In {\it \bibinfo{booktitle}{International Conference on Learning Representations (ICLR2014), CBLS, April 2014}\/}.
\bibitem[{Dwivedi et~al.(2020)Dwivedi, Joshi, Laurent, Bengio \& Bresson}]{dwivedi2020benchmarking}
\bibinfo{author}{Dwivedi, V.~P.}, \bibinfo{author}{Joshi, C.~K.}, \bibinfo{author}{Laurent, T.}, \bibinfo{author}{Bengio, Y.}, \& \bibinfo{author}{Bresson, X.} (\bibinfo{year}{2020}).
\newblock \bibinfo{title}{Benchmarking graph neural networks}.
\newblock {\it \bibinfo{journal}{arXiv preprint arXiv:2003.00982}\/}, .
\bibitem[{Gilmer et~al.(2017)Gilmer, Schoenholz, Riley, Vinyals \& Dahl}]{gilmer2017neural}
\bibinfo{author}{Gilmer, J.}, \bibinfo{author}{Schoenholz, S.~S.}, \bibinfo{author}{Riley, P.~F.}, \bibinfo{author}{Vinyals, O.}, \& \bibinfo{author}{Dahl, G.~E.} (\bibinfo{year}{2017}).
\newblock \bibinfo{title}{Neural message passing for quantum chemistry}.
\newblock In {\it \bibinfo{booktitle}{International conference on machine learning}\/} (pp. \bibinfo{pages}{1263--1272}).
\newblock \bibinfo{organization}{PMLR}.
\bibitem[{Hamilton et~al.(2017)Hamilton, Ying \& Leskovec}]{hamilton2017inductive}
\bibinfo{author}{Hamilton, W.}, \bibinfo{author}{Ying, Z.}, \& \bibinfo{author}{Leskovec, J.} (\bibinfo{year}{2017}).
\newblock \bibinfo{title}{Inductive representation learning on large graphs}.
\newblock In {\it \bibinfo{booktitle}{Advances in neural information processing systems}\/} (pp. \bibinfo{pages}{1024--1034}).
\bibitem[{Hamilton(2020)}]{hamilton2020graph}
\bibinfo{author}{Hamilton, W.~L.} (\bibinfo{year}{2020}).
\newblock \bibinfo{title}{Graph representation learning}.
\newblock {\it \bibinfo{journal}{Synthesis Lectures on Artifical Intelligence and Machine Learning}\/},  {\it \bibinfo{volume}{14}\/}, \bibinfo{pages}{1--159}.
\bibitem[{Hu et~al.(2020)Hu, Fey, Zitnik, Dong, Ren, Liu, Catasta \& Leskovec}]{hu2020open}
\bibinfo{author}{Hu, W.}, \bibinfo{author}{Fey, M.}, \bibinfo{author}{Zitnik, M.}, \bibinfo{author}{Dong, Y.}, \bibinfo{author}{Ren, H.}, \bibinfo{author}{Liu, B.}, \bibinfo{author}{Catasta, M.}, \& \bibinfo{author}{Leskovec, J.} (\bibinfo{year}{2020}).
\newblock \bibinfo{title}{Open graph benchmark: Datasets for machine learning on graphs}.
\newblock {\it \bibinfo{journal}{arXiv preprint arXiv:2005.00687}\/}, .
\bibitem[{Irwin et~al.(2012)Irwin, Sterling, Mysinger, Bolstad \& Coleman}]{irwin2012zinc}
\bibinfo{author}{Irwin, J.~J.}, \bibinfo{author}{Sterling, T.}, \bibinfo{author}{Mysinger, M.~M.}, \bibinfo{author}{Bolstad, E.~S.}, \& \bibinfo{author}{Coleman, R.~G.} (\bibinfo{year}{2012}).
\newblock \bibinfo{title}{Zinc: a free tool to discover chemistry for biology}.
\newblock {\it \bibinfo{journal}{Journal of chemical information and modeling}\/},  {\it \bibinfo{volume}{52}\/}, \bibinfo{pages}{1757--1768}.
\bibitem[{Kingma \& Ba(2014)}]{kingma2014adam}
\bibinfo{author}{Kingma, D.~P.}, \& \bibinfo{author}{Ba, J.} (\bibinfo{year}{2014}).
\newblock \bibinfo{title}{Adam: A method for stochastic optimization}.
\newblock {\it \bibinfo{journal}{arXiv preprint arXiv:1412.6980}\/}, .
\bibitem[{Kipf \& Welling(2017)}]{kipf2017semi}
\bibinfo{author}{Kipf, T.~N.}, \& \bibinfo{author}{Welling, M.} (\bibinfo{year}{2017}).
\newblock \bibinfo{title}{Semi-supervised classification with graph convolutional networks}.
\newblock In {\it \bibinfo{booktitle}{International Conference on Learning Representations (ICLR2017)}\/}.
\bibitem[{Krizhevsky et~al.(2009)Krizhevsky, Hinton et~al.}]{krizhevsky2009learning}
\bibinfo{author}{Krizhevsky, A.}, \bibinfo{author}{Hinton, G.} et~al. (\bibinfo{year}{2009}).
\newblock \bibinfo{title}{Learning multiple layers of features from tiny images}, .
\bibitem[{LeCun et~al.(1998)LeCun, Bottou, Bengio \& Haffner}]{lecun1998gradient}
\bibinfo{author}{LeCun, Y.}, \bibinfo{author}{Bottou, L.}, \bibinfo{author}{Bengio, Y.}, \& \bibinfo{author}{Haffner, P.} (\bibinfo{year}{1998}).
\newblock \bibinfo{title}{Gradient-based learning applied to document recognition}.
\newblock {\it \bibinfo{journal}{Proceedings of the IEEE}\/},  {\it \bibinfo{volume}{86}\/}, \bibinfo{pages}{2278--2324}.
\bibitem[{Li et~al.(2019)Li, Muller, Thabet \& Ghanem}]{li2019deepgcns}
\bibinfo{author}{Li, G.}, \bibinfo{author}{Muller, M.}, \bibinfo{author}{Thabet, A.}, \& \bibinfo{author}{Ghanem, B.} (\bibinfo{year}{2019}).
\newblock \bibinfo{title}{Deepgcns: Can gcns go as deep as cnns?}
\newblock In {\it \bibinfo{booktitle}{Proceedings of the IEEE/CVF international conference on computer vision}\/} (pp. \bibinfo{pages}{9267--9276}).
\bibitem[{Li et~al.(2018)Li, Han \& Wu}]{li2018deeper}
\bibinfo{author}{Li, Q.}, \bibinfo{author}{Han, Z.}, \& \bibinfo{author}{Wu, X.-M.} (\bibinfo{year}{2018}).
\newblock \bibinfo{title}{Deeper insights into graph convolutional networks for semi-supervised learning}.
\newblock In {\it \bibinfo{booktitle}{Proceedings of the AAAI Conference on Artificial Intelligence}\/}.
\newblock volume~\bibinfo{volume}{32}.
\bibitem[{Liu et~al.(2021)Liu, Ding, Jin, Xu, Ma, Liu \& Tang}]{liu2021graph}
\bibinfo{author}{Liu, X.}, \bibinfo{author}{Ding, J.}, \bibinfo{author}{Jin, W.}, \bibinfo{author}{Xu, H.}, \bibinfo{author}{Ma, Y.}, \bibinfo{author}{Liu, Z.}, \& \bibinfo{author}{Tang, J.} (\bibinfo{year}{2021}).
\newblock \bibinfo{title}{Graph neural networks with adaptive residual}.
\newblock {\it \bibinfo{journal}{Advances in Neural Information Processing Systems}\/},  {\it \bibinfo{volume}{34}\/}, \bibinfo{pages}{9720--9733}.
\bibitem[{Monti et~al.(2017)Monti, Boscaini, Masci, Rodola, Svoboda \& Bronstein}]{monti2017geometric}
\bibinfo{author}{Monti, F.}, \bibinfo{author}{Boscaini, D.}, \bibinfo{author}{Masci, J.}, \bibinfo{author}{Rodola, E.}, \bibinfo{author}{Svoboda, J.}, \& \bibinfo{author}{Bronstein, M.~M.} (\bibinfo{year}{2017}).
\newblock \bibinfo{title}{Geometric deep learning on graphs and manifolds using mixture model cnns}.
\newblock In {\it \bibinfo{booktitle}{Proceedings of the IEEE conference on computer vision and pattern recognition}\/} (pp. \bibinfo{pages}{5115--5124}).
\bibitem[{Veli{\v{c}}kovi{\'{c}} et~al.(2018)Veli{\v{c}}kovi{\'{c}}, Cucurull, Casanova, Romero, Li{\`{o}} \& Bengio}]{velickovic2018graph}
\bibinfo{author}{Veli{\v{c}}kovi{\'{c}}, P.}, \bibinfo{author}{Cucurull, G.}, \bibinfo{author}{Casanova, A.}, \bibinfo{author}{Romero, A.}, \bibinfo{author}{Li{\`{o}}, P.}, \& \bibinfo{author}{Bengio, Y.} (\bibinfo{year}{2018}).
\newblock \bibinfo{title}{{Graph Attention Networks}}.
\newblock {\it \bibinfo{journal}{International Conference on Learning Representations}\/}, .
\newblock \bibinfo{note}{Accepted as poster}.
\bibitem[{Wang et~al.(2019)Wang, Yu, Zheng, Gan, Gai, Ye, Li, Zhou, Huang, Ma et~al.}]{wang2019deep}
\bibinfo{author}{Wang, M.}, \bibinfo{author}{Yu, L.}, \bibinfo{author}{Zheng, D.}, \bibinfo{author}{Gan, Q.}, \bibinfo{author}{Gai, Y.}, \bibinfo{author}{Ye, Z.}, \bibinfo{author}{Li, M.}, \bibinfo{author}{Zhou, J.}, \bibinfo{author}{Huang, Q.}, \bibinfo{author}{Ma, C.} et~al. (\bibinfo{year}{2019}).
\newblock \bibinfo{title}{Deep graph library: Towards efficient and scalable deep learning on graphs}.
\newblock {\it \bibinfo{journal}{arXiv preprint arXiv:1909.01315}\/}, .
\bibitem[{Wu et~al.(2019)Wu, Souza, Zhang, Fifty, Yu \& Weinberger}]{wu2019simplifying}
\bibinfo{author}{Wu, F.}, \bibinfo{author}{Souza, A.}, \bibinfo{author}{Zhang, T.}, \bibinfo{author}{Fifty, C.}, \bibinfo{author}{Yu, T.}, \& \bibinfo{author}{Weinberger, K.} (\bibinfo{year}{2019}).
\newblock \bibinfo{title}{Simplifying graph convolutional networks}.
\newblock In {\it \bibinfo{booktitle}{International conference on machine learning}\/} (pp. \bibinfo{pages}{6861--6871}).
\newblock \bibinfo{organization}{PMLR}.
\bibitem[{Xu et~al.(2021)Xu, Zhang, Jegelka \& Kawaguchi}]{xu2021optimization}
\bibinfo{author}{Xu, K.}, \bibinfo{author}{Zhang, M.}, \bibinfo{author}{Jegelka, S.}, \& \bibinfo{author}{Kawaguchi, K.} (\bibinfo{year}{2021}).
\newblock \bibinfo{title}{Optimization of graph neural networks: Implicit acceleration by skip connections and more depth}.
\newblock In {\it \bibinfo{booktitle}{International Conference on Machine Learning}\/} (pp. \bibinfo{pages}{11592--11602}).
\newblock \bibinfo{organization}{PMLR}.
\bibitem[{Zhang et~al.(2023)Zhang, Xia, Zhang \& Xu}]{zhang2023learning}
\bibinfo{author}{Zhang, H.}, \bibinfo{author}{Xia, J.}, \bibinfo{author}{Zhang, G.}, \& \bibinfo{author}{Xu, M.} (\bibinfo{year}{2023}).
\newblock \bibinfo{title}{Learning graph representations through learning and propagating edge features}.
\newblock {\it \bibinfo{journal}{IEEE Transactions on Neural Networks and Learning Systems}\/}, .
\bibitem[{Zhang \& Xu(2024)}]{zhang2024randalign}
\bibinfo{author}{Zhang, H.}, \& \bibinfo{author}{Xu, M.} (\bibinfo{year}{2024}).
\newblock \bibinfo{title}{Randalign: A parameter-free method for regularizing graph convolutional networks}.
\newblock {\it \bibinfo{journal}{arXiv preprint arXiv:2404.09774}\/}, .
\bibitem[{Zhang et~al.(2022{\natexlab{a}})Zhang, Xu, Zhang \& Niwa}]{zhang2022ssfg}
\bibinfo{author}{Zhang, H.}, \bibinfo{author}{Xu, M.}, \bibinfo{author}{Zhang, G.}, \& \bibinfo{author}{Niwa, K.} (\bibinfo{year}{2022}{\natexlab{a}}).
\newblock \bibinfo{title}{Ssfg: Stochastically scaling features and gradients for regularizing graph convolutional networks}.
\newblock {\it \bibinfo{journal}{IEEE Transactions on Neural Networks and Learning Systems}\/}, .
\bibitem[{Zhang et~al.(2022{\natexlab{b}})Zhang, Sheng, Yin, Jiang, Xia, Gao, Yang \& Cui}]{zhang2022model}
\bibinfo{author}{Zhang, W.}, \bibinfo{author}{Sheng, Z.}, \bibinfo{author}{Yin, Z.}, \bibinfo{author}{Jiang, Y.}, \bibinfo{author}{Xia, Y.}, \bibinfo{author}{Gao, J.}, \bibinfo{author}{Yang, Z.}, \& \bibinfo{author}{Cui, B.} (\bibinfo{year}{2022}{\natexlab{b}}).
\newblock \bibinfo{title}{Model degradation ehinders deep graph neural networks}.
\newblock In {\it \bibinfo{booktitle}{Proceedings of the 28th ACM SIGKDD Conference on Knowledge Discovery and Data Mining}\/} (pp. \bibinfo{pages}{2493--2503}).

\end{thebibliography}

\end{document}